\documentclass{article}

\usepackage[final]{corl_2026} %

\makeatletter
\renewcommand{\@conferencelocation}{Austin, Texas, USA}
\makeatother

\usepackage{amsmath}
\usepackage{amssymb}
\usepackage{booktabs}
\usepackage{graphicx}
\usepackage{algorithm}
\usepackage{algpseudocode}
\usepackage{xcolor}
\usepackage{pifont}
\usepackage{subcaption}
\usepackage{wrapfig}
\usepackage{placeins}
\usepackage{enumitem}
\definecolor{shoesred}{HTML}{D62728}
\newcommand{\cmark}{\ding{51}}
\newcommand{\xmark}{\ding{55}}

\title{SHOES: Sim-to-real Hardware-adaptation via On-policy
Expansion of Skills for Legged Robots}

\author{
  Kota Tsuruhara$^{1,2}$ \quad Tsung-Yuan Chen$^{2}$ \quad Kento Kawaharazuka$^{2}$ \\[0.3em]
  \normalfont
  $^{1}$Tokyo University of Agriculture and Technology \quad
  $^{2}$The University of Tokyo \\
  \texttt{\{trhrkota, chentsungyuan\}@g.ecc.u-tokyo.ac.jp} \\
  \texttt{kawaharazuka@jsk.imi.i.u-tokyo.ac.jp} \\
}

\begin{document}
\maketitle

\begin{figure}[!h]
  \centering
  \includegraphics[width=0.78\linewidth]{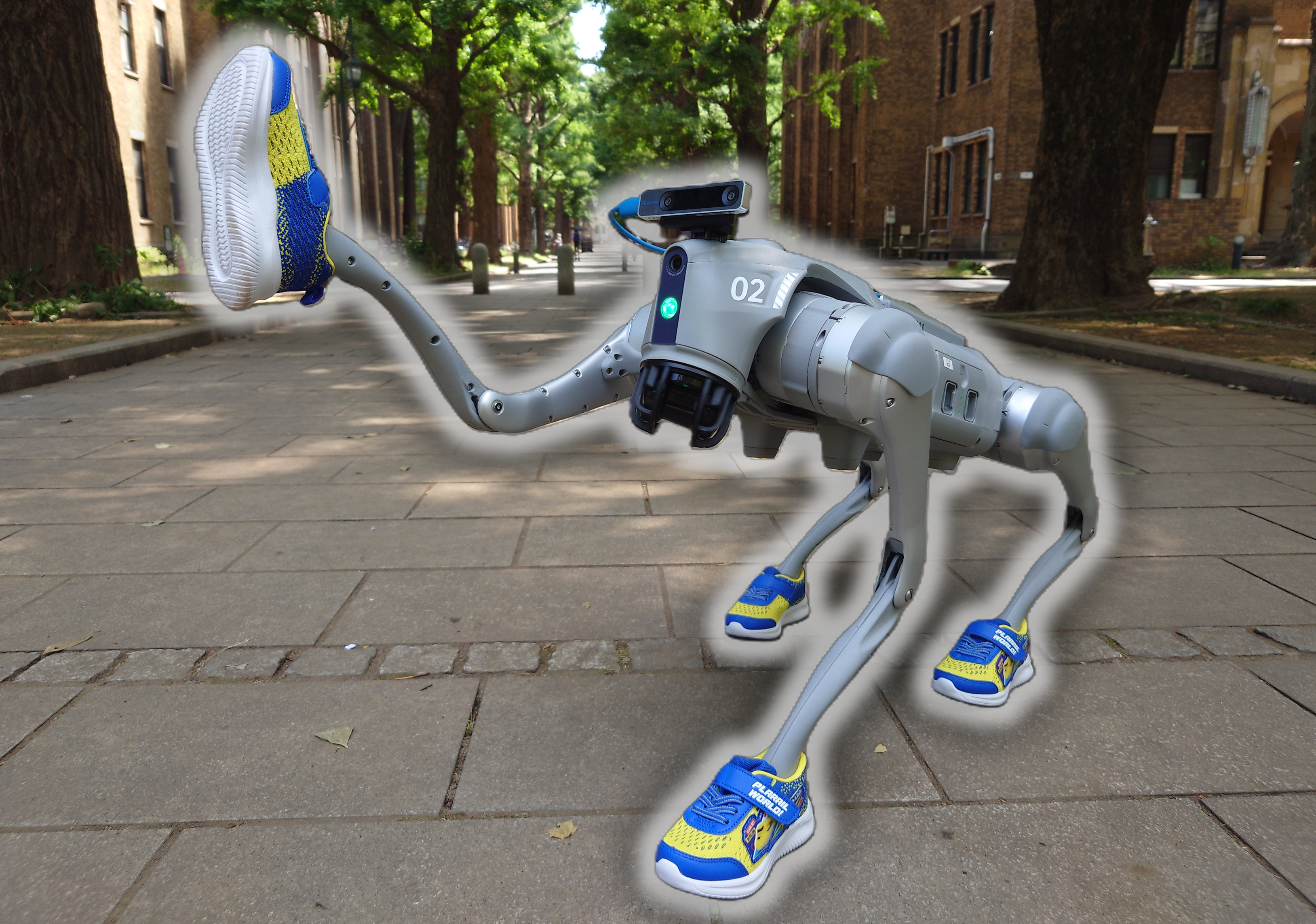}
  \caption{A Unitree~Go2 wearing shoes.}
  \label{fig:teaser}
\end{figure}

\begin{abstract}
  Humans expand their locomotion capability by wearing appropriate footwear,
  such as hiking boots for mountains or skates for ice. Footwear changes the
  ground-contact interface and enables traversal of otherwise difficult
  terrain. In contrast, locomotion controllers are typically trained for a
  fixed morphology, and their performance can degrade when shoes alter foot
  geometry, mass, and friction. We propose \textbf{SHOES}, a sample-efficient
  framework for adapting locomotion policies to footwear-induced morphology
  changes. SHOES freezes a simulation-trained nominal policy and fine-tunes
  only lightweight, identity-initialized modulation modules on the robot,
  preserving the initial behavior while enabling efficient hardware
  adaptation. We also introduce \textbf{SafeBox}, a workspace-aware waypoint
  sampler that constrains online adaptation within a user-defined workspace.
  The simulation results with a shoe-equipped Unitree~Go2 show that SHOES
  substantially recovers locomotion performance after footwear-induced
  morphology changes using limited additional adaptation. Under the same
  adaptation budget, SHOES outperforms FiLM-only, Linear Adapter-only, and
  residual-RL baselines. Real-world experiments on a Unitree~Go2 fitted with
  human shoes further evaluate the effectiveness of SHOES for hardware
  deployment.
\end{abstract}


\section{Introduction}
\label{sec:intro}

\begin{wrapfigure}{r}{0.40\linewidth}
  \centering
  \includegraphics[width=\linewidth]{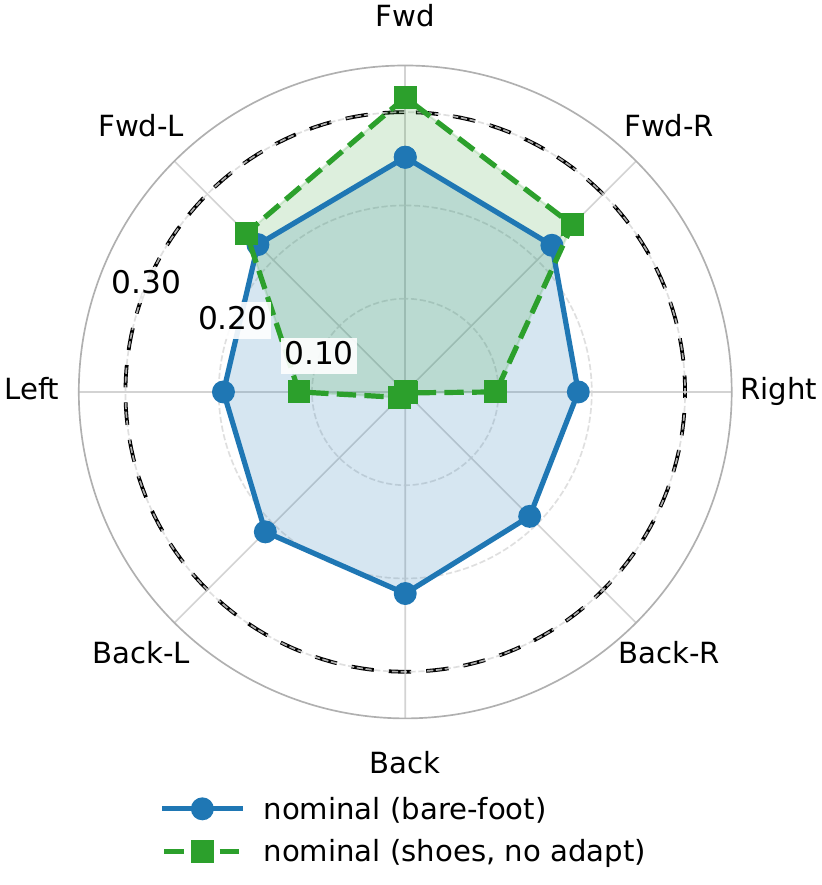}
  \caption{
    \textbf{Bare-foot nominal degrades with shoes attached.}
    Commanded-direction velocity at the $0.3$~m/s setpoint for the
    same bare-foot nominal policy, executed on bare feet (blue) and
    after attaching shoes (green); the dashed ring marks the target.
    The morphology shift induced by shoes particularly degrades
    backward-command tracking.
  }
  \label{fig:nominal_lowspeed}
\end{wrapfigure}

  Humans expand their locomotion capabilities by changing footwear, from
  hiking boots to skates. Legged robots could similarly benefit from shoes
  that improve traction or protect floors and hardware
  (Figure~\ref{fig:teaser}). However, footwear changes the robot's morphology
  and its interaction with the ground. It alters the
  foot-contact geometry, frictional behavior, compliance, mass
  distribution, and effective leg geometry. A policy trained for the
  bare-foot robot can therefore degrade when the same controller is
  executed with shoes attached. Figure~\ref{fig:nominal_lowspeed}
  shows this failure mode on a Unitree~Go2: attaching shoes changes the
  robot's locomotion characteristics, and the bare-foot nominal policy
  becomes particularly poor at tracking backward commands.

  Domain randomization (DR) improves robustness to deployment-time mismatch
  by varying simulation parameters during training~\citep{tobin2017dr, peng2018simdr, chebotar2019closing}.
  Footwear-aware DR requires specifying variations in shoe geometry,
  contact properties, mass, and inertia in advance. Although this can
  improve zero-shot robustness, broader randomization increases modeling
  and training costs and can induce conservative behavior.

  Another option is to fine-tune the full locomotion policy directly on
  hardware~\citep{smith2022keeplearning}. Here, we instead preserve the
  simulation-trained controller and learn only the correction required by
  the attached shoes.

  This work develops such a post-training adaptation approach for
  footwear-induced morphology changes. Our main contributions are:

  \begin{itemize}[leftmargin=12pt,itemsep=2pt]
    \item We propose \textbf{SHOES} (Sim-to-real Hardware-adaptation
    via On-policy Expansion of Skills), a lightweight framework for
    footwear-induced morphology adaptation that preserves a
    simulation-trained nominal actor and fine-tunes only identity-initialized
    FiLM and Linear Adapter modules.

    \item We introduce SafeBox, a workspace-aware waypoint sampler that
    supports on-robot adaptation by collecting data under a consistent
    command interface in simulation and on hardware.

    \item We evaluate SHOES in simulation and real-robot experiments,
    showing that it substantially recovers bare-foot tracking performance
    on shoe-equipped Unitree~Go2 robots and outperforms FiLM-only, Linear
    Adapter-only, and residual-RL baselines.
  \end{itemize}

\section{Related Work}
\label{sec:related}

  \paragraph{Domain randomization for legged sim-to-real.}
  DR is a standard recipe for improving sim-to-real robustness in
  legged locomotion by training policies over randomized dynamics,
  contact, terrain, and sensing parameters. It has been widely
  deployed in quadrupedal sim-to-real RL under diverse deployment
  conditions
  \citep{tan2018simtoreal, hwangbo2019anymal, lee2020anymal, kumar2021rma, miki2022wildanymal, ji2022concurrent}
  and for more advanced quadruped capabilities
  \citep{margolis2022rapid, margolis2022walkthese, shao2022gaittransition, feng2022genloco, wu2026robogauge},
  and similarly underpins bipedal and humanoid sim-to-real RL across
  diverse platforms and skills
  \citep{yu2019simtorealbiped, li2021robustparameterized, radosavovic2024humanoid, zhuang2024humanoidparkour}.
  A long line of Cassie work has further developed DR-trained
  skill-specific controllers
  \citep{siekmann2020memory, siekmann2021periodic, siekmann2021blindstairs, duan2022footstep},
  following an earlier non-DR iterative-design baseline
  \citep{xie2019cassie}.
  \citet{rudin2022minutes} made this recipe practical through massively
  parallel training, while \citet{liu2025locoformer} extends it to
  procedurally generated robots with long-context in-context adaptation.
  SHOES is complementary to DR: footwear-aware DR requires a predefined
  distribution of contact properties, whereas SHOES adapts to the
  encountered condition after deployment and can also be applied to a
  DR-trained policy.

  \paragraph{Real-world RL and sim-to-online fine-tuning.}
  Smith et al.~\citep{smith2022walkinpark} demonstrate SAC-based reinforcement
  learning~\citep{haarnoja2018sac} from scratch on
  a real quadruped in 20 minutes, and Wu et al.~\citep{wu2023daydreamer} learn
  world models on hardware. Without a competent sim-trained initial policy,
  learning locomotion directly on hardware can be sample-inefficient and
  require careful safety supervision during early exploration~\citep{ha2020learning}.
  Smith et al.~\citep{smith2022keeplearning} fine-tune the full policy using
  off-policy RL and a learned recovery policy. SHOES instead starts from a
  nominal policy trained with on-policy PPO, a standard choice in quadruped
  locomotion. Full-policy fine-tuning also requires a separate full-sized
  checkpoint per footwear, whereas SHOES reuses the same nominal policy and
  switches only small footwear-specific modules.
  \citet{as2026simtoonline} provide a systematic study of sim-to-online
  RL design choices on real robots, including data retention and delayed
  actor updates. Related lines of work align simulation and real-world
  physics for agile humanoid whole-body skills \citep{he2025asap} and
  develop rapid online adaptive control for robust legged locomotion
  \citep{lyu2024rl2ac}. Their focus, however, is on algorithmic or
  physics-alignment design choices rather than identity-initialized
  adaptation modules around a frozen nominal policy.

  \paragraph{FiLM-based fine-tuning on hardware.}
  FiLM~\citep{perez2018film} provides a general-purpose
  feature-wise affine conditioning layer. The closest hardware-side
  fine-tuning method to ours is RTR~\citep{hu2025rtr}, which
  fine-tunes a FiLM-conditioned latent on a humanoid robot while
  using a teacher robot arm for safety. SHOES differs in three
  key aspects: (i)~modulation is conditioned directly on the policy
  observation rather than on a learned dynamics latent; (ii)~the
  FiLM modulation is bounded and is attached to a nominal policy
  trained without FiLM, with identity initialization preserving
  the nominal actor exactly at the start of fine-tuning; and
  (iii)~SHOES separates the correction into state-dependent
  hidden-layer modulation and an action-space Linear Adapter, a
  design choice that is validated by the ablation study in
  Section~\ref{sec:ablation}.

\section{Preliminaries}
\label{sec:prelim}

  We model footwear adaptation as a two-stage reinforcement
  learning problem on a fully observable MDP
  $(\mathcal{S}, \mathcal{A}, P, r, \gamma)$. A policy
  $\pi_\theta(a\mid o)$ outputs a Gaussian over actions $a$
  conditioned on the proprioceptive observation $o$. Concrete
  observation, action, and control-rate specifications are
  deferred to Appendix~\ref{app:hyperparams}.

  \paragraph{Stage~1 (pre-deploy, sim).}
  We first train a nominal policy $\pi_\text{nom}$ in
  simulation, solving
  \begin{equation}
    \pi_\text{nom} \;=\; \arg\max_{\pi}\;
    \mathbb{E}_{\tau \sim \pi}\!
      \left[\sum_t r_\text{bare}(s_t, a_t)\right],
    \label{eq:stage1}
  \end{equation}
  until convergence with a large simulation budget. The Stage-1
  reward $r_\text{bare}$ is the standard quadruped velocity
  recipe (Appendix~\ref{app:rewards}).

  \paragraph{Stage~2 (post-deploy, online).}
  The robot is then equipped with shoes that alter the
  foot-ground contact dynamics; under this morphology shift the
  zero-shot performance of $\pi_\text{nom}$ may degrade.
  Stage~2 freezes $\pi_\text{nom}$ and trains a small adapter
  parameterized by $\theta$ on top, subject to a budget
  $N_\text{adapt}$:
  \begin{equation}
    \theta^{\star} \;=\; \arg\max_{\theta}\;
    \mathbb{E}_{\tau \sim \pi_{\theta}}\!
      \left[\sum_t r_\text{shoe}(s_t, a_t)\right],
    \quad
    \text{s.t. } \pi_{\theta}(\cdot)\big|_{\theta=0} = \pi_\text{nom}(\cdot),
    \ \
    N_\text{steps} \le N_\text{adapt}.
    \label{eq:stage2}
  \end{equation}
  The identity-at-initialization constraint
  $\pi_\theta|_{\theta=0} = \pi_\text{nom}$ ensures that Stage~2
  starts regression-free with respect to the sim-trained base.
  The Stage-2 reward $r_\text{shoe}$ retains only the
  velocity-tracking terms of $r_\text{bare}$ with amplified
  weights (Appendix~\ref{app:rewards}): stable locomotion is already
  learned in Stage~1, so Stage~2 optimizes only velocity tracking.

\begin{figure}[t]
  \centering
  \includegraphics[width=\linewidth]{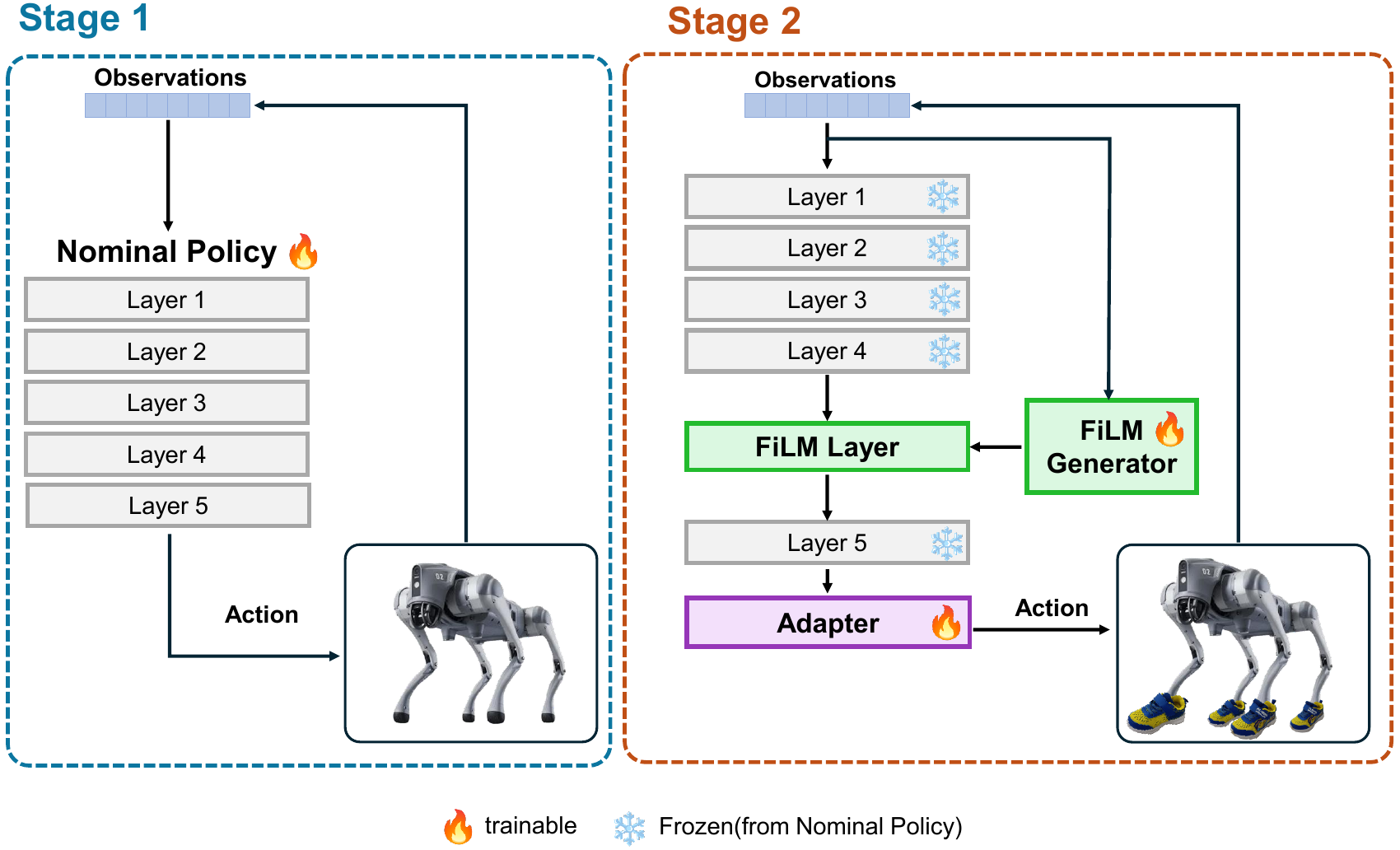}
  \caption{
    \textbf{SHOES two-stage training pipeline.} Stage~1 trains a
    nominal PPO actor-critic on the bare-foot Go2 in simulation
    with $4096$ parallel envs; the resulting $\pi_\text{nom}$
    becomes the frozen base for Stage~2.
    Stage~2 wraps the frozen nominal MLP (gray, ${\sim}190$k params)
    with a trainable FiLM generator and a Linear Adapter
    (coral, ${\sim}5.6$k params); PPO online-fine-tunes the
    trainable modules on the shoe-equipped Go2 with the SafeBox
    sampler at $1$ env $\times\,30$k env-steps.
  }
  \label{fig:arch}
\end{figure}

\section{Method}
\label{sec:method}

  \paragraph{Overview.}
  SHOES adapts a simulation-trained locomotion policy to new
  footwear conditions while keeping the nominal policy interface
  unchanged. Starting from a Stage-1 nominal policy, SHOES freezes
  the nominal actor and observation normalizer, and fine-tunes
  only lightweight, identity-initialized adaptation modules in
  Stage~2. The framework has two architectural components: a
  bounded FiLM generator paired with a Linear Adapter for
  behavior-preserving policy adaptation (Sec.~\ref{sec:arch}),
  and SafeBox for workspace-aware waypoint sampling during
  on-robot adaptation (Sec.~\ref{sec:safebox}). Stage~2 fine-tunes
  these modules with on-policy PPO~\citep{schulman2017ppo}; full pseudocode and
  hyper-parameters are deferred to
  Appendices~\ref{app:algorithm} and~\ref{app:hyperparams}. An
  overview of SHOES is shown in Fig.~\ref{fig:arch}.

\subsection{Behavior-Preserving Policy Adaptation}
\label{sec:arch}

  SHOES adapts the nominal policy by wrapping the frozen actor with
  lightweight trainable modules, while preserving the original policy
  interface. These modules are initialized so that, at the beginning of
  Stage~2, the wrapped actor is exactly equivalent to the Stage-1
  nominal policy. Figure~\ref{fig:arch} shows this wrapped actor.
  Let the frozen nominal actor be decomposed into a feature extractor
  $f_\text{body}: \mathbb{R}^{d_o} \to \mathbb{R}^{d_h}$ and an
  output head
  $f_\text{head}: \mathbb{R}^{d_h} \to \mathbb{R}^{d_a}$, where
  $d_o$, $d_h$, and $d_a$ denote the observation, hidden, and
  action dimensions, respectively:
  \begin{equation}
    h \;=\; f_\text{body}(o), \qquad
    \mu_\text{nom} \;=\; f_\text{head}(h).
    \label{eq:nominal-decomp}
  \end{equation}
  Around this frozen computation, we insert two trainable
  modules:
  \begin{equation}
    h' \;=\; \mathrm{FiLM}_{\phi}(h, o), \qquad
    \mu \;=\; A_{\psi}\bigl(f_\text{head}(h')\bigr).
    \label{eq:shoes-decomp}
  \end{equation}
  Here, $f_\text{body}$ and $f_\text{head}$ are frozen, while
  $\phi$ (the FiLM generator) and $\psi$ (the Linear Adapter)
  are trained online with PPO. The two trainable modules are
  instantiated as follows.

  \paragraph{Bounded FiLM modulation.} A small FiLM generator
  $g_\phi: \mathbb{R}^{d_o}\to\mathbb{R}^{2 d_h}$ consumes the
  (normalized) observation $o$ and outputs scale and shift parameters
  $(\gamma_\text{raw}, \beta_\text{raw}) \in
  \mathbb{R}^{d_h} \times \mathbb{R}^{d_h}$. The hidden
  representation is modulated as
  \begin{equation}
    h' \;=\; \underbrace{\bigl(1 + \alpha\,\tanh(\gamma_\text{raw})\bigr)}_{\text{multiplicative scale}}
        \odot\, h
        \;+\;
        \underbrace{\alpha\,\tanh(\beta_\text{raw})}_{\text{additive shift}},
    \qquad \alpha = 0.5.
    \label{eq:film}
  \end{equation}
  We initialize the final layer of $g_\phi$ to zero so that
  $\gamma_\text{raw} = \beta_\text{raw} = 0$ at $\theta = 0$, making
  the FiLM modulation the identity map $h' = h$ at initialization.
  The $(1+\alpha\tanh)$ form bounds the multiplicative gain to
  $[1-\alpha, 1+\alpha]$ and the additive shift to $[-\alpha, \alpha]$;
  with $\alpha = 0.5$ this restricts modulation to a
  nominal-anchored neighborhood under the limited on-robot
  adaptation budget.

  \paragraph{Linear Adapter.} An affine map
  $A_\psi: \mathbb{R}^{d_a}\to\mathbb{R}^{d_a}$ post-processes the
  action mean:
  \begin{equation}
    \mu \;=\; W_A \,\bigl[f_\text{head}(h')\bigr] \,+\, b_A.
    \label{eq:adapter}
  \end{equation}
  We initialize $W_A = I_{d_a}$ and $b_A = 0$ so that $A_\psi$
  is the identity map $\mu = f_\text{head}(h')$ at initialization.

  The Linear Adapter does not see $o$; it captures static
  action-space offsets, such as systematic leg-height biases induced by
  footwear, while FiLM provides observation-conditioned modulation of the
  hidden representation. Together, they separate static output-space
  corrections from dynamic hidden-state modulation.

  Combining \eqref{eq:film} and \eqref{eq:adapter}, the actor performs
  a single forward pass per environment step. At initialization,
  $h' = h$ and $W_A = I$, so $\mu = \mu_\text{nom}$ exactly.

\subsection{Workspace-Aware Waypoint Sampling (SafeBox)}
\label{sec:safebox}

  Random velocity commands can drive a robot outside the available
  workspace during online adaptation. SafeBox instead generates commands
  toward waypoints sampled within a user-defined workspace, without changing
  the policy architecture, action space, or training objective.

  \paragraph{Workspace-aware waypoint sampling.}
  SafeBox is parameterized by a user-defined rectangular workspace
  $\mathcal{B}=[-a,a]\times[-b,b]$ in a local world frame. At each
  resampling event, it samples a waypoint
  $\mathbf{t}=(t_x,t_y)$ uniformly from $\mathcal{B}$, a cruise speed
  $v_\mathrm{cap}$, and a yaw-rate command $\omega_z$:
  \begin{equation}
    \mathbf{t} \sim \mathcal{U}(\mathcal{B}), \qquad
    v_\mathrm{cap}
    \sim
    \mathcal{U}
    \bigl(v_\mathrm{cap}^{\min}, v_\mathrm{cap}^{\max}\bigr),
    \qquad
    \omega_z
    \sim
    \mathcal{U}
    \bigl(-\omega_{\max}, \omega_{\max}\bigr).
    \label{eq:safebox-sampling}
  \end{equation}
  Here $v_\mathrm{cap}$ is the (scalar) magnitude of the world-frame
  velocity command used in Eq.~\eqref{eq:safebox-vel}, and the per-axis
  limits $v_x^{\max}, v_y^{\max}$ in Eq.~\eqref{eq:safebox-cmd} act on
  the body-frame components after projection. We reuse the forward limit
  as the cruise-speed upper bound, $v_\mathrm{cap}^{\max} := v_x^{\max}$,
  since forward motion dominates the training distribution; this couples
  the two scalars by convention rather than by definition. The sampled
  waypoint and yaw-rate command are held until the robot reaches a small
  neighborhood of the waypoint or a resampling timer expires.

  \paragraph{Closed-loop velocity command generation.}
  Given the current planar position $\mathbf{p}=(x_w,y_w)$ and yaw
  $\psi$, SafeBox first computes a world-frame velocity that points
  toward the active waypoint:
  \begin{equation}
    d = \|\mathbf{t}-\mathbf{p}\|,
    \qquad
    \mathbf{v}^{w}
    =
    \min\!\bigl(v_\mathrm{cap}, k_p d\bigr)
    \frac{\mathbf{t}-\mathbf{p}}{\max(d,\delta)} .
    \label{eq:safebox-vel}
  \end{equation}
  Here, $k_p$ is a proportional gain and $\delta$ is a small constant for
  numerical stability. The command magnitude saturates far from the
  waypoint and decreases smoothly near the target. The world-frame
  velocity is rotated into the body frame, clipped per axis, and combined
  with the sampled yaw-rate command:
  \begin{equation}
    \begin{pmatrix} v_x^{\mathrm{body}} \\ v_y^{\mathrm{body}} \end{pmatrix}
    = R(-\psi)\,\mathbf{v}^{w},
    \qquad
    \mathbf{v}^{\mathrm{cmd}}
    =
    \begin{pmatrix}
      \mathrm{clip}\!\left(v_x^{\mathrm{body}},\,-v_x^{\max},\,v_x^{\max}\right) \\[2pt]
      \mathrm{clip}\!\left(v_y^{\mathrm{body}},\,-v_y^{\max},\,v_y^{\max}\right) \\[2pt]
      \omega_z
    \end{pmatrix}.
    \label{eq:safebox-cmd}
  \end{equation}
  The policy receives $\mathbf{v}^{\mathrm{cmd}}$ through the same command
  channels as in Stage~1 and does not observe the robot position, sampled
  waypoint, or workspace bounds. Since each waypoint is sampled inside
  $\mathcal{B}$ and each planar command points toward the active
  waypoint, SafeBox discourages long open-loop drifts while preserving the
  policy interface. Implementation details and parameter values are
  provided in Appendix~\ref{app:safebox}.

\section{Experiments}
\label{sec:exp}

  We evaluate SHOES on a Unitree Go2 in simulation using the
  MuJoCo-Warp engine through the mjlab environment
  framework~\citep{zakka2026mjlab}. The bare-foot robot defines the
  reference morphology and the shoe-equipped robot defines the
  post-deployment morphology shift to be adapted. Unless noted
  otherwise, all Stage-2 methods start from the same frozen nominal
  checkpoint, use the SafeBox command sampler, and receive the same
  single-environment adaptation budget of
  $N_\text{adapt} = 30{,}000$ environment steps.

  Performance is measured on a $24$-cell velocity-tracking benchmark
  formed by crossing $3$ speed setpoints with $8$ directions. The
  primary figure of merit, Bare-Foot Parity (BFP),
  aggregates the per-cell ratio of the policy's commanded-direction
  velocity to the bare-foot nominal's: $\text{BFP} = 100\,\%$
  indicates parity with the bare-foot reference in the commanded
  direction, and lower scores mean the policy tracks more slowly than
  bare-foot. Per-cell measurement details, the exact formula, and the
  multi-seed protocol are in Appendix~\ref{app:eval}.

\begin{figure}[!htbp]
    \centering
    \includegraphics[width=\linewidth]{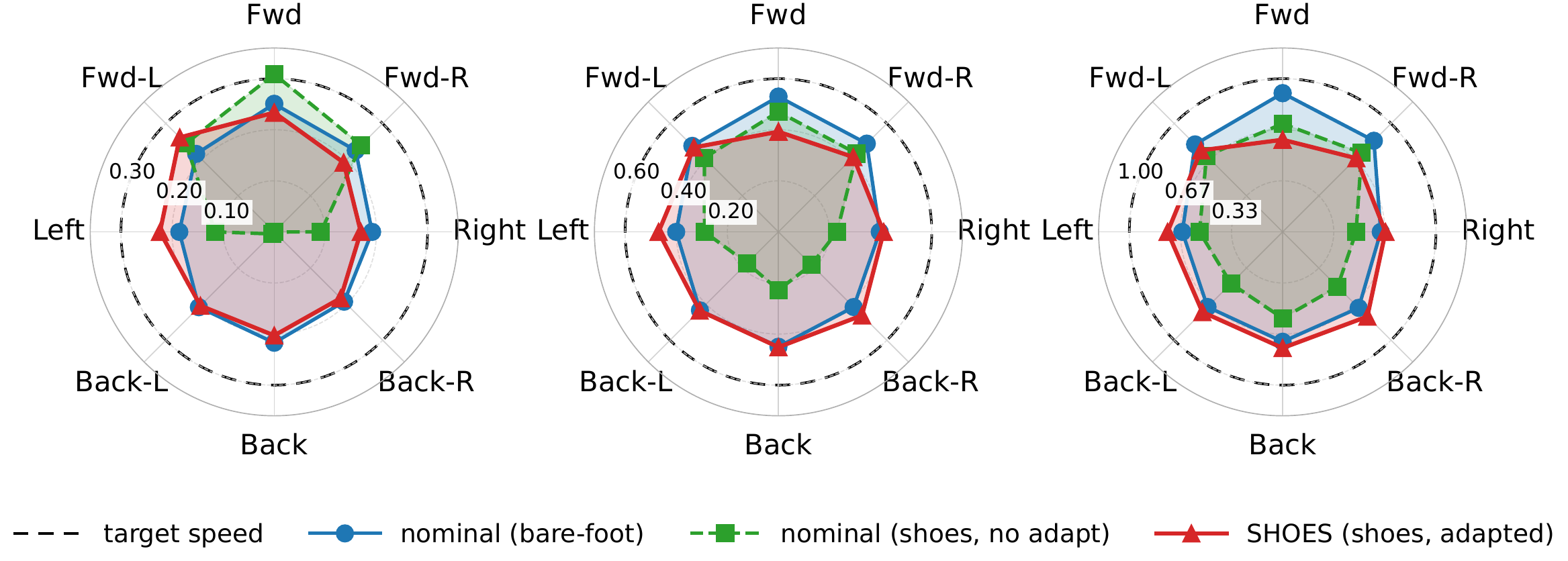}
    \caption{
      \textbf{SHOES recovers bare-foot tracking across speeds and directions.}
      Per-direction tracked velocity at the three command speeds
      ($0.3, 0.6, 1.0$~m/s). SHOES (red, best seed from
      Appendix~\ref{app:main_table}) closely matches the bare-foot
      nominal reference (blue), while the no-adapt nominal on shoes
      (green) tracks well inside it across all directions and speeds.
    }
    \label{fig:filmadapter_solo}
  \end{figure}

  \paragraph{Q1. Can SHOES recover from footwear-induced degradation?}
  \label{sec:main}
  We evaluate whether SHOES can recover the tracking performance lost
  when the frozen bare-foot nominal is executed with shoes attached.
  Without adaptation, the nominal drops to $\text{BFP} = 67.0\,\%$;
  after a 30k-step Stage-2 fine-tune, SHOES reaches
  $93.5 \pm 3.9\,\%$ over six seeds, recovering most of the loss.
  Figure~\ref{fig:filmadapter_solo} provides the per-direction
  breakdown: across all three speed setpoints, the no-adapt nominal
  sits well inside the bare-foot nominal, whereas SHOES closely
  matches it, indicating broad recovery across the 24-cell tracking
  benchmark.

  \FloatBarrier
  \newpage
  \begin{wraptable}{r}{0.56\linewidth}
    \centering
    \footnotesize
    \setlength{\tabcolsep}{3pt}
    \caption{\textbf{Component ablation} ($30$k-step budget).
    $\Delta$BFP is the gain over the no-adapt nominal (first row).}
    \label{tab:ablation_comp}
    \begin{tabular}{ccrrr}
      \toprule
      \textbf{FiLM} & \textbf{Adapter} & \multicolumn{1}{c}{\textbf{Trainable}} & \multicolumn{1}{c}{\textbf{BFP $\uparrow$}} & \multicolumn{1}{c}{\textbf{$\Delta$BFP $\uparrow$}} \\
      \midrule
      \xmark & \xmark & $0$    & $67.0$               & $0.0$ \\
      \xmark & \cmark & $0.2$k & $65.1 \pm 8.9$       & $-1.9$ \\
      \cmark & \xmark & $5.4$k & $78.7 \pm 10.8$      & $+11.7$ \\
      \textcolor{shoesred}{\cmark} & \textcolor{shoesred}{\cmark} & \textcolor{shoesred}{$\mathbf{5.6}$\textbf{k}} & \textcolor{shoesred}{$\mathbf{93.5 \pm 3.9}$} & \textcolor{shoesred}{$\mathbf{+26.5}$} \\
      \bottomrule
    \end{tabular}
  \end{wraptable}
  \paragraph{Q2. Is the decomposition of hidden-feature and action-space corrections necessary?}\label{sec:ablation}
  SHOES separates footwear adaptation into state-dependent hidden-layer
  modulation and an action-space Linear Adapter. We test whether this
  separation is necessary by ablating each component under the same
  30k-step adaptation budget. Table~\ref{tab:ablation_comp} shows that
  the Linear Adapter alone fails to improve over the no-adapt nominal,
  indicating that a static action-space correction cannot by itself
  compensate for the shoe-induced shift. FiLM-only recovers a
  meaningful portion of the gap but still falls short of the bare-foot
  reference. Combining the two components closes most of the gap,
  recovering $+26.5$\,pp of Bare-Foot Parity over the no-adapt
  nominal. This result supports the core SHOES design: under a limited
  on-robot-equivalent budget, footwear adaptation is better captured
  by combining state-dependent feature modulation with a lightweight
  correction in the action space around the frozen nominal actor.

  \begin{wrapfigure}{r}{0.45\linewidth}
    \centering
    \includegraphics[width=\linewidth]{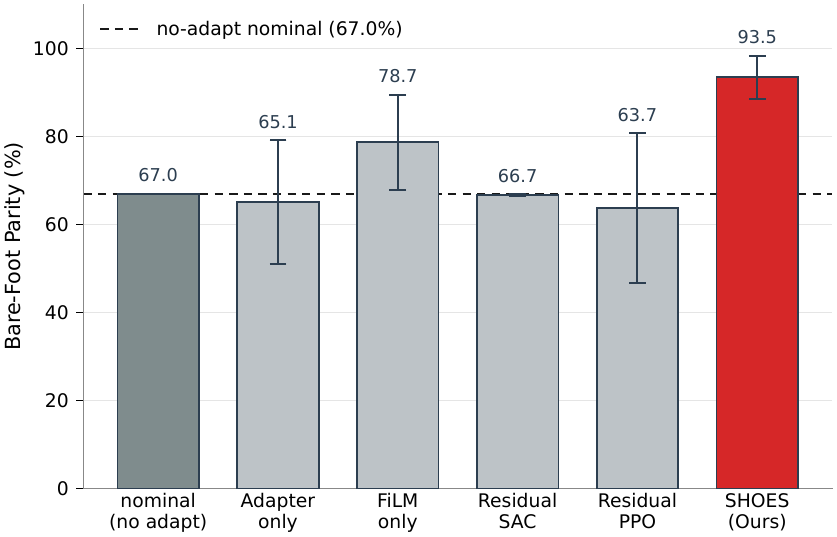}
    \caption{
      \textbf{Method comparison (flat shoes).}
      Bare-Foot Parity (\%) mean~$\pm$~std over six seeds.
      Dashed line: no-adapt nominal.
    }
    \label{fig:method_bar}
  \end{wrapfigure}
  \paragraph{Q3. Which post-training adaptation method is most effective?}
  We compared SHOES with alternative post-training adaptation baselines
  under the same 30k-step budget. Figure~\ref{fig:method_bar} shows
  that the Linear Adapter, Residual SAC, and Residual PPO baselines
  all remain near the no-adapt nominal, failing to make meaningful
  progress within the on-robot-equivalent budget. FiLM-only is the
  strongest baseline but still falls short of the bare-foot reference.
  Only the full SHOES model substantially closes the gap, giving the
  largest recovery among all methods. The comparison indicates that
  effective post-training adaptation is not achieved by adding
  residual or single-module capacity alone; under the limited
  adaptation budget, the identity-initialized combination of
  hidden-feature modulation and action-space correction is what
  separates SHOES from the baselines. We additionally compared SHOES with
  full-policy PPO fine-tuning under the same
  30k-step adaptation budget. Averaged over the four additional footwear
  variants evaluated in Q4, SHOES achieves a BFP of $107.3 \pm 3.3$\%,
  compared with $97.8 \pm 2.3$\% for full-policy PPO fine-tuning.

\begin{figure}[!htbp]
    \centering
    \includegraphics[width=0.6\linewidth]{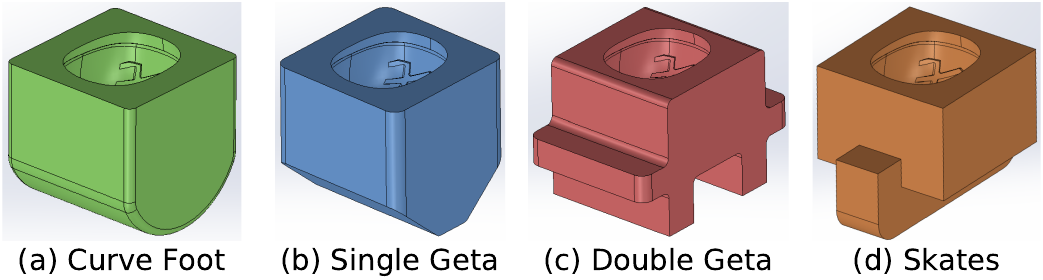}
    \caption{Simulated footwear variants: (a) Curve Foot, (b) Single Geta,
    (c) Double Geta, and (d) Skates.}
    \label{fig:shoe_variants}
  \end{figure}

  \paragraph{Q4. Does SHOES adapt beyond flat shoes?}
  \label{sec:variants}
  We added simulated footwear variants that alter contact geometry and added
  mass beyond the flat shoe (Figure~\ref{fig:shoe_variants}). For each footwear
  variant, SHOES starts from the same nominal checkpoint and adapts for 30k
  environment steps. SHOES recovers performance across all of them, as summarized in
  Table~\ref{tab:variants}. Since BFP measures only velocity along the
  commanded direction, we additionally report fall rate, lateral velocity
  error, yaw-rate tracking error, and action-rate cost in
  Table~\ref{tab:metrics}, including evaluations with nonzero yaw-rate
  commands. SHOES improves commanded-velocity tracking and stability,
  while the mean action-rate cost increases modestly from $0.725$ to $0.790$.
  Indeed, Table~\ref{tab:metrics} shows that optimizing only
  velocity tracking in Stage~2 also reduces the fall rate.

  \begin{table}[!htbp]
    \centering
    \footnotesize
    \setlength{\tabcolsep}{3pt}
    \begin{minipage}[t]{0.44\linewidth}
      \centering
      \caption{Velocity-tracking recovery across simulated footwear variants.}
      \label{tab:variants}
      \begin{tabular}{lcc}
        \toprule
        Footwear variant & \begin{tabular}[c]{@{}c@{}}Nominal\\(w/o adapt.)\end{tabular} & SHOES (ours) \\
        \midrule
        Curve Foot & 36.4\% & $115.0 \pm 5.8$\% \\
        Single Geta & 36.8\% & $113.5 \pm 6.5$\% \\
        Double Geta & 17.7\% & $103.6 \pm 12.7$\% \\
        Skates & 38.6\% & $97.2 \pm 5.1$\% \\
        \bottomrule
      \end{tabular}
    \end{minipage}
    \hfill
    \begin{minipage}[t]{0.54\linewidth}
      \centering
      \caption{Stability and motion-quality metrics averaged over the four
      additional footwear variants.}
      \label{tab:metrics}
      \begin{tabular}{lcc}
        \toprule
        Metric & \begin{tabular}[c]{@{}c@{}}Nominal\\(w/o adapt.)\end{tabular} & SHOES (ours) \\
        \midrule
        Fall rate & 55.4\% & $6.3 \pm 4.6$\% \\
        Lateral error (m/s) & 0.148 & $0.080 \pm 0.005$ \\
        Yaw-rate error (rad/s) & 0.106 & $0.070 \pm 0.005$ \\
        Action-rate cost & 0.725 & $0.790 \pm 0.043$ \\
        \bottomrule
      \end{tabular}
    \end{minipage}
  \end{table}

\begin{figure}[!htbp]
    \centering
    \begin{subfigure}[c]{0.48\linewidth}
      \centering
      \includegraphics[width=\linewidth]{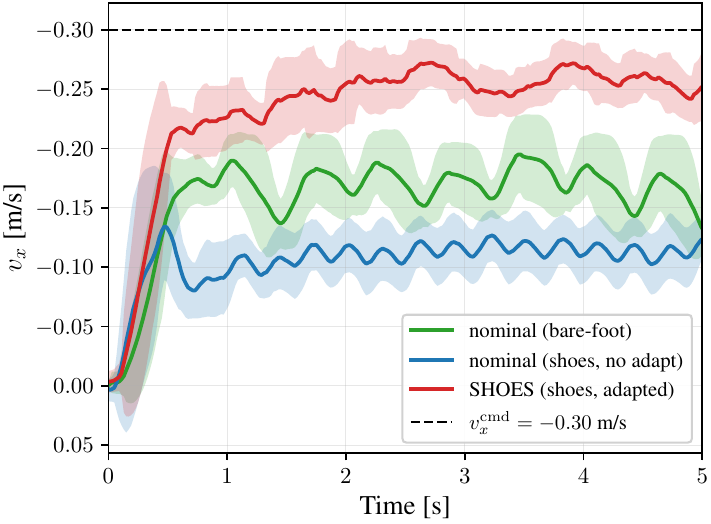}
      \caption{Body-frame $v_x$ tracking (rolling mean $\pm 1\sigma$,
      $21$-sample window) at $v_x^{\text{cmd}} = -0.30$~m/s.}
      \label{fig:linvel}
    \end{subfigure}
    \hfill
    \begin{subfigure}[c]{0.48\linewidth}
      \centering
      \setlength{\fboxsep}{0pt}\setlength{\fboxrule}{0.5pt}%
      \fbox{\includegraphics[width=\dimexpr\linewidth-2\fboxrule\relax]{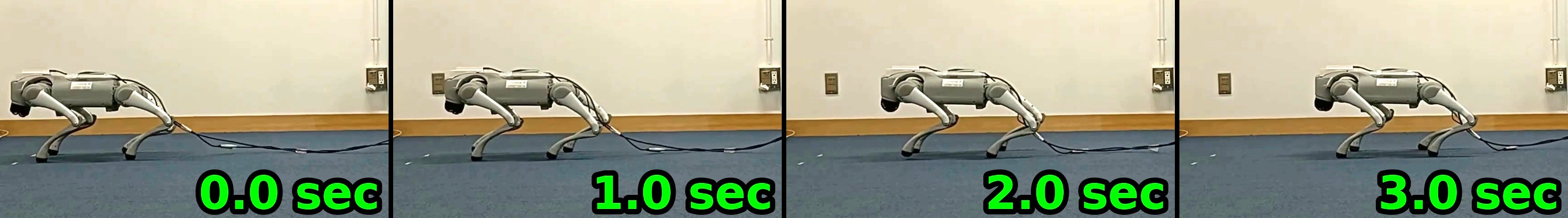}}\\[0.2em]
      {\footnotesize nominal (bare-foot)}\\[0.5em]
      \fbox{\includegraphics[width=\dimexpr\linewidth-2\fboxrule\relax]{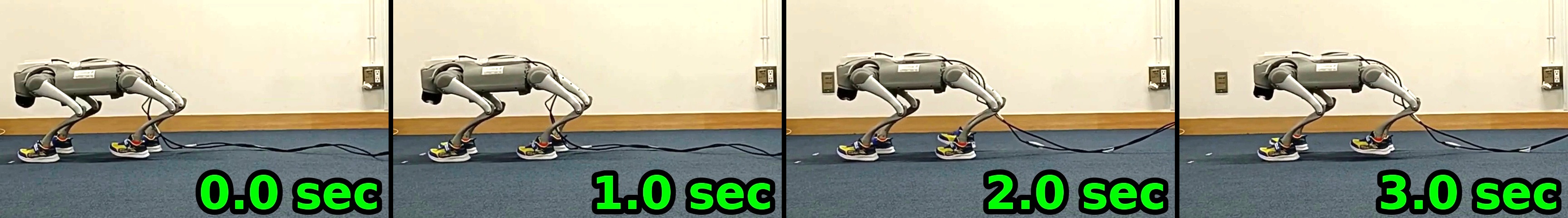}}\\[0.2em]
      {\footnotesize nominal (shoes, no adapt)}\\[0.5em]
      \fbox{\includegraphics[width=\dimexpr\linewidth-2\fboxrule\relax]{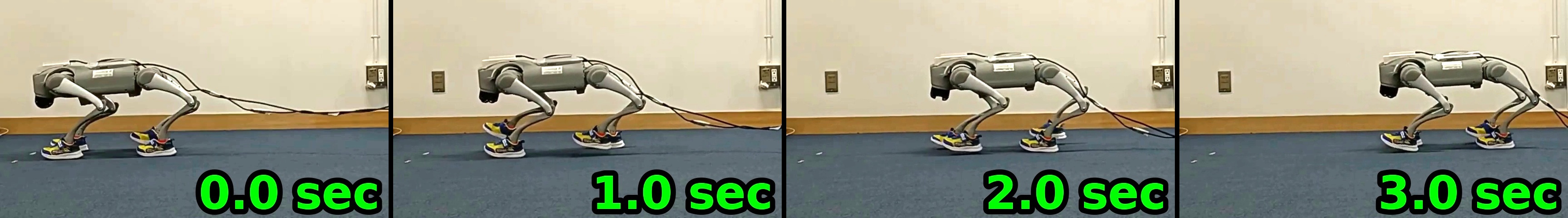}}\\[0.2em]
      {\footnotesize SHOES (shoes, adapted)}
      \caption{On-robot snapshots.}
      \label{fig:real_snapshots}
    \end{subfigure}
    \caption{
      \textbf{SHOES recovers backward-velocity tracking on hardware.}
      Backward setpoint $v_x^{\text{cmd}} = -0.30$~m/s on the physical
      Unitree~Go2 with flat shoes.
      (a)~SHOES (red) substantially improves backward tracking over
      the no-adapt nominal on shoes (blue); the bare-foot nominal
      (green) is shown for reference (black dashed line: command).
      (b)~Snapshots of the robot in the three settings overlaid in
      the left panel.
    }
    \label{fig:onrobot}
  \end{figure}

  \FloatBarrier
  \paragraph{Q5. Can the same adaptation interface be deployed on hardware?}
  \label{sec:realrobot}
  We finally examined whether the Stage-2 adaptation transfers to the
  physical Unitree~Go2 with shoes attached
  (Figure~\ref{fig:onrobot}). The policy runs at $50$~Hz on top of
  the $500$~Hz onboard PD loop. We evaluated the backward setpoint
  $v_x^{\text{cmd}} = -0.30$~m/s, which exhibited the largest no-adapt
  gap in the simulation radar chart (Figure~\ref{fig:filmadapter_solo}).
  Figure~\ref{fig:linvel} reports the body-frame $v_x$ over a
  constant-command window.

  The no-adapt nominal policy with shoes attached (blue) tracked at
  $v_x \approx -0.10 \pm 0.03$~m/s, corresponding to only $33\,\%$ of
  the commanded velocity. In contrast, SHOES (red) reached
  $v_x \approx -0.25 \pm 0.03$~m/s, recovering $83\,\%$ of the command
  and reducing the steady-state tracking error from $0.20$ to
  $0.05$~m/s. The bare-foot nominal policy (green), evaluated on the
  same hardware as the reference morphology, tracked at
  $v_x \approx -0.15 \pm 0.04$~m/s, or $50\,\%$ of the command. This
  remaining gap reflects the sim-to-real discrepancy already present
  without the footwear-induced morphology shift.

  Notably, SHOES exceeded the bare-foot nominal reference on the same
  hardware in this backward-velocity test. This result suggests that
  on-robot Stage-2 fine-tuning not only compensates for the
  footwear-induced morphology change, but can also reduce part of the
  pre-existing sim-to-real gap of the nominal policy.
  Figure~\ref{fig:real_snapshots} qualitatively illustrates this
  difference on the physical robot: attaching shoes to the no-adapt
  nominal visibly slows locomotion, whereas Stage-2 fine-tuning with
  SHOES restores backward speed on the same hardware.

  Throughout the $30$k-step Stage-2 fine-tuning process, the SafeBox
  sampler constrained exploration safely: no episode required emergency
  intervention, no boundary violation was observed, and data collection
  proceeded without hardware damage. These results confirm improved
  tracking over the unadapted shoe-equipped policy, with SHOES additionally
  surpassing the bare-foot nominal reference in this backward-velocity test.

\begin{figure}[!htbp]
    \centering
    \includegraphics[width=0.85\linewidth]{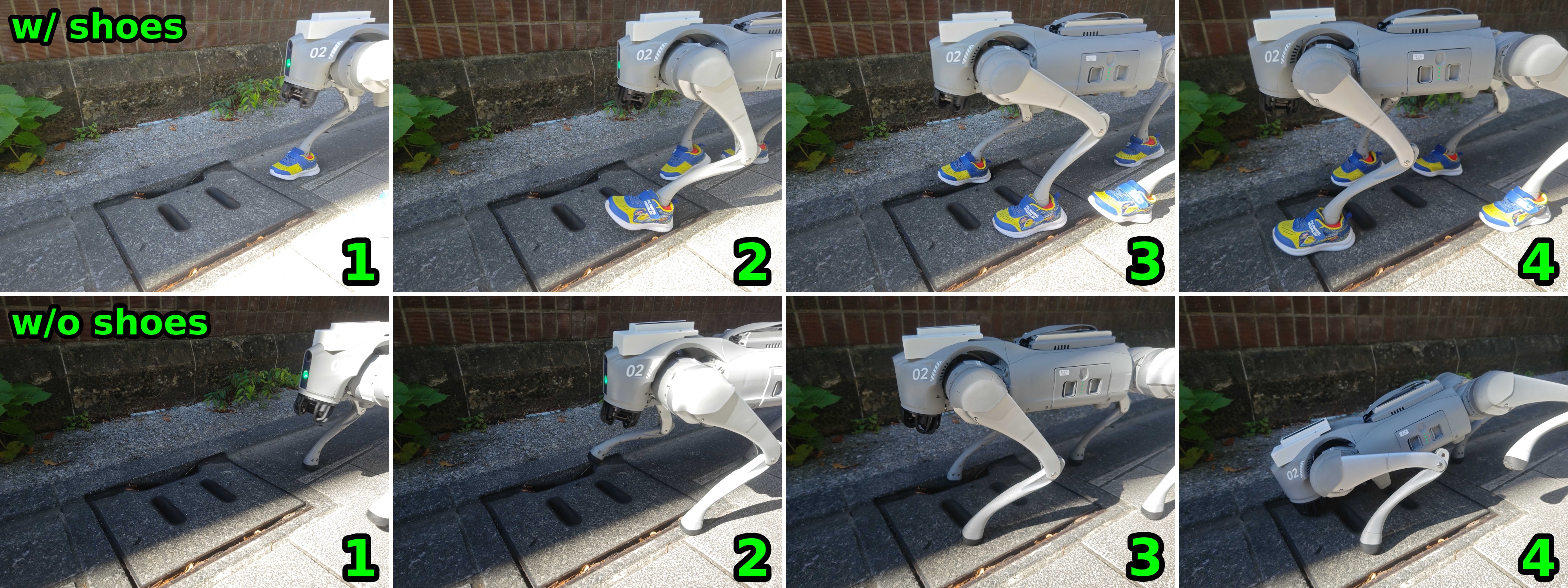}
    \caption{Real-robot traversal over a slotted surface. With shoes, the
    robot crosses the openings (top); without shoes, it becomes trapped
    (bottom).}
    \label{fig:slot}
  \end{figure}

  \FloatBarrier
  \paragraph{Q6. Does footwear enable traversal beyond basic protection?}
  \label{sec:slot}
  We added a real-robot experiment on a slotted surface
  (Figure~\ref{fig:slot}). Without shoes, a foot drops into an opening and
  the robot becomes trapped, whereas the shoe-equipped robot adapted with
  SHOES traverses the surface. Task-specific footwear enables traversal of
  terrain that is inaccessible to the bare-foot nominal robot, and SHOES
  adapts the locomotion policy to the new footwear condition.

  \FloatBarrier

\section{Limitations}
\label{sec:limitations}

  SafeBox enables on-robot Stage-2 fine-tuning by generating
  exploratory commands within a user-defined workspace. This
  reduces the need for continuous operator intervention to keep
  rollouts inside the workspace. However, the current system does
  not integrate physical resets into the online adaptation loop.
  Consequently, SHOES is not yet practical for footwear conditions
  that frequently induce falls or other reset-triggering failures.
  In such cases, training must be suspended until an operator
  manually rights the robot and restarts the episode. For footwear causing
  frequent failures, SHOES can be combined with a learned self-righting
  policy or a safety-tether reset triggered upon fall detection.
  Incorporating fall detection, autonomous stand-up recovery, and
  reset-aware episode management could extend online adaptation to more
  severe footwear-induced morphology shifts.

\section{Conclusion}
\label{sec:conclusion}

  We presented SHOES, a sample-efficient framework for adapting
  locomotion policies to footwear-induced morphology changes.
  SHOES freezes a simulation-trained nominal policy and
  fine-tunes only lightweight, identity-initialized FiLM and
  Linear Adapter modules. This lets fine-tuning start from the
  nominal policy behavior and gradually adapt it to the
  shoe-equipped robot. SafeBox supports online adaptation by
  keeping exploratory rollouts within a user-defined workspace.
  On a shoe-equipped Unitree~Go2, SHOES reached a bare-foot parity
  of $93.5 \pm 3.9\,\%$ using only
  $30$k environment steps. These results suggest a practical
  route for adapting deployed controllers to post-deployment
  hardware changes.

\clearpage
\acknowledgments{We thank the anonymous reviewers and the area chair for
their constructive feedback, which led to the additional footwear
variants, complementary metrics, and the slotted-terrain experiment in the
camera-ready version. This work was supported by JST Moonshot R\&D Program
Grant Number JPMJMS263D.}

\bibliography{corl}  %

\appendix

\section*{Appendix}

\section{Architecture and Hyper-parameters}
\label{app:hyperparams}

  \paragraph{Architecture.} We instantiate Section~\ref{sec:arch} with
  $d_o = 47$, $d_h = 128$, and $d_a = 12$. The $47$-d observation $o$ stacks
  base angular velocity ($3$), projected gravity ($3$), velocity command
  ($3$), per-joint positions and velocities ($12+12$), the previous action
  ($12$), and a $\sin\!/\cos$ phase signal ($2$). The $12$-d action specifies
  joint-position offsets executed by a $500$~Hz onboard PD loop. The frozen
  nominal MLP has hidden widths $(512, 256, 128)$ with ELU activations. The
  FiLM generator $g_\phi$ is
  $\text{Linear}(47{\to}16){\to}\text{ELU}{\to}\text{Linear}(16{\to}16){\to}\text{ELU}{\to}\text{Linear}(16{\to}256)$
  with the final layer zero-initialized. The Linear Adapter is a single
  $d_a\!\times\!d_a$ affine layer initialized to identity. The trainable
  actor totals $5{,}560$ parameters ($5{,}392$ FiLM + $156$ Linear Adapter +
  $12$ action log-std); the warm-started critic has ${\sim}190$k parameters.

  \paragraph{Training.} Stage~2 uses PPO (Table~\ref{tab:hyperparams}) with
  $\text{num\_steps\_per\_env}\!=\!256$, $K\!=\!3$ epochs per update, and an
  adaptive KL learning rate ($\eta_0\!=\!3\!\times\!10^{-5}$, target KL
  $0.005$). With $\text{num\_envs}\!=\!1$ and
  $N_\text{adapt}\!=\!30{,}000$ we run $117$ PPO iterations; the SafeBox
  sampler supplies velocity commands throughout. At identity initialization,
  a unit test verifies that the first actor forward pass is bit-identical to
  $\pi_\text{nom}$ to $10^{-5}$ absolute action error. The FiLM bound
  $\alpha=0.5$ and generator hidden dimensions $(16,16)$ are selected from
  the sweep in Appendix~\ref{app:sweeps}. Stage~2 fine-tuning completes in
  ${\sim}10$~min on a single RTX~4070; Stage~1 nominal training uses $4096$
  parallel envs for $30$M env-steps on an RTX~5080.

  \begin{table}[h]
    \centering
    \small
    \caption{Stage-2 PPO hyper-parameters.}
    \label{tab:hyperparams}
    \begin{tabular}{lr}
      \toprule
      Parameter & Value \\
      \midrule
      Parallel environments & $1$ \\
      Steps per environment per rollout & $256$ \\
      PPO epochs per update & $3$ \\
      Mini-batches per epoch & $4$ \\
      Total PPO iterations & $117$ \\
      Total environment steps & $30{,}000$ \\
      Clip parameter & $0.1$ \\
      Initial learning rate & $3\!\times\!10^{-5}$ \\
      Adaptive KL target & $0.005$ \\
      Discount $\gamma$ & $0.99$ \\
      GAE $\lambda$ & $0.95$ \\
      FiLM bound $\alpha$ & $0.5$ \\
      Generator hidden dims & $(16, 16)$ \\
      Optimizer & Adam (default $\beta$) \\
      \bottomrule
    \end{tabular}
  \end{table}

\section{Reward Specification}
\label{app:rewards}

  Both stages optimize $R(\tau) = \sum_t \sum_i w_i \, r_i(s_t, a_t)$.

  \paragraph{Stage~1 (bare-foot nominal training).}
  Combines velocity tracking, posture, and gait-shaping terms following
  the standard quadruped recipe (Table~\ref{tab:rewards-stage1}).

  \begin{table}[h]
    \centering
    \small
    \caption{Stage-1 rewards (bare-foot nominal training).
    $\mathbf{v}^\text{cmd}_{xy}, \omega_z^\text{cmd}$: commanded
    linear/yaw velocity (body frame);
    $\mathbf{v}_{xy}, v_z, \boldsymbol{\omega}_{xy}$: base linear
    and angular velocities;
    $\mathbf{g}^\text{proj}_{xy}$: projected gravity (body horizontal);
    $\boldsymbol{\theta},
    \ddot{\boldsymbol{\theta}}$: joint positions and accelerations;
    $\boldsymbol{\theta}^\text{default}$: default joint pose;
    $\mathbf{L}_\text{root}$: root angular momentum;
    $p_z^i, \mathbf{v}^i_{xy}, \mathbf{F}^i$: foot $i$ height, xy
    velocity (world), contact force;
    $c_i\in\{0,1\}$: stance indicator; $\eta_i\in\{0,1\}$:
    first-contact (landing) indicator;
    $\theta_j^\text{low}, \theta_j^\text{up}$: soft joint position limits;
    ``terminated'' covers fall (base tilt $>70^\circ$) and non-foot contact
    ($>10$~N); $T_\text{phase} = 0.6$~s, $p_z^\text{tgt} = 0.10$~m,
    $\sigma_\text{lin}^2 = 0.25$, $\sigma_\text{ang}^2 = 0.5$.}
    \label{tab:rewards-stage1}
    \begin{tabular}{llr}
      \toprule
      Reward & Equation $r_i$ & Weight $w_i$ \\
      \midrule
      Lin. velocity tracking & $\exp\!\bigl(-[\lVert \mathbf{v}^\text{cmd}_{xy} - \mathbf{v}_{xy} \rVert^2 + 2 v_z^2] / \sigma_\text{lin}^2\bigr)$ & $+1.0$ \\
      Ang. velocity tracking & $\exp\!\bigl(-[(\omega_z^\text{cmd} - \omega_z)^2 + 0.05\lVert\boldsymbol{\omega}_{xy}\rVert^2] / \sigma_\text{ang}^2\bigr)$ & $+1.0$ \\
      Body orientation & $\lVert \mathbf{g}^\text{proj}_{xy} \rVert^2$ & $-1.0$ \\
      Posture & $\exp\!\bigl(-\tfrac{1}{|\mathcal{J}|}\sum_{j}(\theta_j - \theta_j^\text{default})^2 / \sigma_{\text{pose},j}(\mathbf{v}^\text{cmd})^2\bigr)$ & $+1.0$ \\
      Body angular velocity & $\lVert \boldsymbol{\omega}_{xy} \rVert^2$ & $-0.05$ \\
      Angular momentum & $\lVert \mathbf{L}_\text{root} \rVert^2$ & $-0.025$ \\
      Termination & $\mathbb{1}\{\text{terminated}\}$ & $-200$ \\
      Joint accelerations & $\lVert \ddot{\boldsymbol{\theta}} \rVert^2$ & $-2.5\!\times\!10^{-7}$ \\
      Joint position limits & $\sum_j \bigl[\mathrm{ReLU}(\theta_j^\text{low} - \theta_j) + \mathrm{ReLU}(\theta_j - \theta_j^\text{up})\bigr]$ & $-10$ \\
      Action rate & $\lVert \mathbf{a}_t - \mathbf{a}_{t-1} \rVert^2$ & $-0.05$ \\
      Foot gait & $\tfrac{1}{4}\sum_i \mathbb{1}\{c_i = s_i(t)\}$ & $+0.5$ \\
      Foot clearance & $\sum_i |p_z^i - p_z^\text{tgt}|\cdot \lVert \mathbf{v}^i_{xy} \rVert$ & $-1.0$ \\
      Foot slip & $\sum_i \lVert \mathbf{v}^i_{xy} \rVert^2 \cdot c_i$ & $-0.25$ \\
      Soft landing & $\sum_i \lVert\mathbf{F}^i\rVert\cdot \eta_i$ & $-10^{-3}$ \\
      Stand still & $\lVert \boldsymbol{\theta} - \boldsymbol{\theta}^\text{default} \rVert^2 \cdot \mathbb{1}\{u^\text{cmd} \le \varepsilon\}$ & $-1.0$ \\
      \bottomrule
    \end{tabular}
  \end{table}

  The trot stance schedule is
  $s_i(t) = \mathbb{1}\!\bigl\{\bigl((t/T_\text{phase}) + \phi_i\bigr)\!\bmod 1 < 0.56\bigr\}$
  with diagonal offsets $\phi = (0, 0.5, 0.5, 0)$. The posture standard
  deviation $\sigma_{\text{pose},j}$ depends on the joint and command
  magnitude: standing
  $(\sigma_\text{hip}, \sigma_\text{thigh}, \sigma_\text{calf}) = (0.05, 0.10, 0.15)$
  when $u^\text{cmd} := \lVert \mathbf{v}^\text{cmd}_{xy}\rVert + |\omega_z^\text{cmd}| < 0.1$,
  walking $(0.15, 0.35, 0.50)$ otherwise. The foot-related terms
  (gait, clearance, slip) and stand\_still are gated by the same $u^\text{cmd}$
  threshold $\varepsilon = 0.1$; soft\_landing uses $0.05$.

  \paragraph{Stage~2 (SHOES online adaptation).}
  Every Stage-1 shaping term is set to zero in Stage~2
  (Table~\ref{tab:rewards-stage2}): the shaping rewards
  are already baked into the frozen nominal MLP, and keeping non-zero
  weights empirically collapses the policy into a ``stand-still / walk
  conservatively'' local optimum. The linear-velocity term uses an
  adaptive standard deviation,
  \begin{equation}
    \sigma_\text{lin}(s) \;=\;
    \max\!\bigl(\sigma_\text{min},\; k \cdot
      \lVert \mathbf{v}^\text{cmd}_{xy} \rVert\bigr),
    \qquad
    \sigma_\text{min} = 0.05,\; k = 0.3.
    \label{eq:adaptive-std}
  \end{equation}

  \begin{table}[h]
    \centering
    \small
    \caption{Stage-2 rewards. $\sigma_\text{lin}(s)$ from
    Eq.~\eqref{eq:adaptive-std}, $\sigma_\text{ang} = 0.25$.}
    \label{tab:rewards-stage2}
    \begin{tabular}{llr}
      \toprule
      Reward & Equation $r_i$ & Weight $w_i$ \\
      \midrule
      Lin. velocity tracking (adaptive) & $\exp\!\bigl(-[\lVert \mathbf{v}^\text{cmd}_{xy} - \mathbf{v}_{xy} \rVert^2 + 2 v_z^2] / \sigma_\text{lin}(s)^2\bigr)$ & $+10.0$ \\
      Ang. velocity tracking & $\exp\!\bigl(-[(\omega_z^\text{cmd} - \omega_z)^2 + 0.05\lVert\boldsymbol{\omega}_{xy}\rVert^2] / \sigma_\text{ang}^2\bigr)$ & $+2.0$ \\
      \bottomrule
    \end{tabular}
  \end{table}

\section{SHOES Stage-2 Procedure}
\label{app:algorithm}

  Algorithm~\ref{alg:shoes} summarizes the SHOES Stage~2
  adaptation loop used in both simulation and on-robot runs.

  \begin{algorithm}[t]
    \caption{SHOES on-robot adaptation.}
    \label{alg:shoes}
    \begin{algorithmic}[1]
      \State Train nominal policy $\pi_\text{nom}$ in simulation on the bare-foot robot (Stage~1).
      \State Freeze observation normalizer and nominal actor parameters.
      \State Initialize FiLM generator final layer to zero and Linear Adapter to identity ($W_A=I_{d_a}, b_A=0$).
      \State Warm-start critic from the nominal critic.
      \For{each PPO iteration}
        \For{each environment step}
          \State Sample velocity command $(v_x, v_y, \omega_z)_t$ with SafeBox.
          \State Obtain robot pose and observation $o_t$.
          \State Compute $h_t = f_\text{body}(o_t)$ (frozen).
          \State Apply bounded FiLM: $h'_t = \mathrm{FiLM}_{\phi}(h_t, o_t)$.
          \State Compute action mean $\mu_t = A_{\psi}\bigl(f_\text{head}(h'_t)\bigr)$.
          \State Sample $a_t \sim \mathcal{N}(\mu_t, \exp(\log\sigma)^2)$ and execute the PD target.
          \State Compute reward $r_t$ and store transition.
        \EndFor
        \State Update FiLM ($\phi$), Linear Adapter ($\psi$), log-standard deviation ($\log\sigma$), and critic with PPO.
      \EndFor
    \end{algorithmic}
  \end{algorithm}

\section{SafeBox Default Parameters}
\label{app:safebox}

  Defaults (Table~\ref{tab:safebox-params}) are identical in simulation and
  on hardware. The waypoint is drawn uniformly from the box and held until
  the robot enters the target-reached radius $\epsilon$, or on hardware
  until the hold-time timer expires. Eq.~\eqref{eq:safebox-vel} is projected
  to the body frame and clipped per Eq.~\eqref{eq:safebox-cmd} at every
  $50$~Hz tick using the latest yaw, so yaw drift bends the command smoothly
  rather than producing a step change.

  \begin{table}[h]
    \centering
    \small
    \caption{SafeBox sampler defaults.}
    \label{tab:safebox-params}
    \begin{tabular}{lr}
      \toprule
      Parameter & Value \\
      \midrule
      Box half-width $a$ (x-axis) & $1.5$~m \\
      Box half-width $b$ (y-axis) & $1.0$~m \\
      Cruise-speed lower bound $v_\text{cap}^{\min}$ & $0.1$~m/s \\
      Cruise-speed upper bound $v_\text{cap}^{\max} (= v_x^{\max})$ / body-frame $|v_x|$ clip $v_x^{\max}$ & $1.0$~m/s \\
      Body-frame $|v_y|$ clip $v_y^{\max}$ & $0.5$~m/s \\
      Max $|\omega_z|$ ($\omega_\text{max}$) & $0.5$~rad/s \\
      Position gain $k_p$ & $2.0$~s$^{-1}$ \\
      Target-reached threshold $\epsilon$ & $0.1$~m \\
      Hold-time window (hardware) & $[3, 8]$~s \\
      \bottomrule
    \end{tabular}
  \end{table}

\section{Evaluation Protocol}
\label{app:eval}

  Performance is measured on a $24$-cell velocity-tracking benchmark
  formed by crossing $3$ speed setpoints ($0.3, 0.6, 1.0$~m/s) with
  $8$ directions ($45^{\circ}$ increments). Each cell is evaluated
  with $\text{num\_envs}=16$ parallel environments, $100$ warmup
  steps, and $T = 200$ measurement steps. For each cell $c$ with
  commanded direction $\hat{\mathbf{c}}_c \in \mathbb{R}^2$, we
  record the commanded-direction body-frame velocity, averaged over
  time and environments and clipped at zero:
  \begin{equation}
    v^c \;=\; \max\!\Bigl(0,\;\overline{\mathbf{v}_{xy}} \cdot \hat{\mathbf{c}}_c\Bigr),
    \qquad
    \overline{\mathbf{v}_{xy}} \;=\; \frac{1}{T}\sum_{t=1}^{T}\mathbf{v}_{xy}(t).
    \label{eq:vc-def}
  \end{equation}
  The primary figure of merit is the Bare-Foot Parity (BFP), the
  mean per-cell ratio against the bare-foot nominal,
  \begin{equation}
    \text{BFP} \;=\;
    \frac{1}{24}\sum_{c=1}^{24}
      \frac{v^{c}_\text{policy}}{v^{c}_\text{bare}}
      \times 100\,\%.
    \label{eq:pct-bare}
  \end{equation}
  A score of $\text{BFP} = 100\,\%$ indicates parity with the bare-foot
  reference in the commanded direction; lower scores mean the policy tracks
  more slowly. Note that BFP measures only the commanded-direction
  component:
  lateral drift and yaw error contribute only insofar as they reduce it. The
  main-table and component-ablation rows aggregate six seeds
  (mean~$\pm$~std and best); the sweeps in Appendix~\ref{app:sweeps} use a
  single seed per setting.

\section{Per-Method Results, Full Numbers}
\label{app:main_table}

  Table~\ref{tab:main} expands Figure~\ref{fig:method_bar} with the best-seed
  column. Each row is six seeds at $30$k env-steps with the SafeBox sampler
  from a common nominal checkpoint. The first row is the frozen nominal
  evaluated on the shoe-equipped robot.

  \begin{table}[h]
    \centering
    \small
    \caption{Method comparison on flat shoes. Mean~$\pm$~std and best
    Bare-Foot Parity (BFP) over six seeds.}
    \label{tab:main}
    \begin{tabular}{lrrr}
      \toprule
      \textbf{Method} & \multicolumn{1}{c}{\textbf{Trainable}} & \multicolumn{1}{c}{\textbf{BFP $\uparrow$}} & \multicolumn{1}{c}{\textbf{Best $\uparrow$}} \\
      \midrule
      nominal (no adapt) & $0$ & $67.0$ & --- \\
      Adapter-only & $0.2$k & $65.1 \pm 8.9$ & $76.1$ \\
      FiLM-only & $5.4$k & $78.7 \pm 10.8$ & $90.3$ \\
      Residual SAC & ${\sim}38$k & $66.7 \pm 0.2$ & $66.8$ \\
      Residual PPO & ${\sim}38$k & $63.7 \pm 8.3$ & $81.8$ \\
      \textcolor{shoesred}{\textbf{SHOES (Ours)}} & \textcolor{shoesred}{$\mathbf{5.6}$\textbf{k}} & \textcolor{shoesred}{$\mathbf{93.5 \pm 3.9}$} & \textcolor{shoesred}{$\mathbf{96.9}$} \\
      \bottomrule
    \end{tabular}
  \end{table}

\section{Real-Robot Protocol}
\label{app:real_protocol}

  A $50$~Hz policy thread reads the latest IMU and T265 pose, builds the
  same $47$-d observation as in simulation, and publishes the next PD target;
  a $500$~Hz onboard PD thread executes it at the joint level. Reward is
  computed online from T265 linear velocity (rotated into the body frame via
  the IMU quaternion) and IMU yaw rate, using the Stage-2 reward
  (Appendix~\ref{app:rewards}). SafeBox supplies the velocity command, with
  the T265 providing world-frame pose. An emergency-stop button on the
  wireless controller triggers a damping-mode hand-off independent of the
  policy pipeline.

\section{FiLM Bound Width and Generator Capacity Sweeps}
\label{app:sweeps}

  Single-seed trends justifying the FiLM hyper-parameter choices.

  \paragraph{Bound width $\alpha$.} The multiplicative scale is
  $1 + \alpha\,\tanh(\gamma_\text{raw}) \in [1-\alpha, 1+\alpha]$. Unbounded
  FiLM ($\alpha\!\to\!\infty$, no $\tanh$ saturation) collapses to a
  fall-avoidance shortcut around $67\,\%$; $\alpha\!=\!0.3$ is too tight and
  trails $\alpha\!=\!0.5$; $\alpha\!=\!1.0$ trends back toward the unbounded
  shortcut. We use $\alpha = 0.5$ throughout.

  \paragraph{Generator hidden dimension.} For the generator
  $\text{Linear}(47\!\to\!h)\!\to\!\text{ELU}\!\to\!\text{Linear}(h\!\to\!h)\!\to\!\text{ELU}\!\to\!\text{Linear}(h\!\to\!256)$
  we use $h\!=\!16$. $h\!=\!8$ underfits the shoe-induced shift;
  $h\!\in\!\{32, 64\}$ overfits and trends toward the unbounded shortcut.

\end{document}